\documentclass[sigconf]{acmart}

\AtBeginDocument{%
  \providecommand\BibTeX{{%
    \normalfont B\kern-0.5em{\scshape i\kern-0.25em b}\kern-0.8em\TeX}}}

\acmConference[Conference DLG-KDD '22]{KDD 2022 Workshop on Deep Learning on Graphs: Methods and Applications}{August 15,
  2022}{Washington, D.C.}

\usepackage{algorithm}
\usepackage{algorithmic}
\usepackage{amsmath}
\DeclareMathOperator*{\argmax}{arg\,max}
\begin{document}

\title{Subgraph Frequency Distribution Estimation \\ using Graph Neural Networks}

%
\author{Zhongren Chen}
\authornote{Both authors contributed equally to this research.}
\email{zhongren.chen@stanford.edu}
\affiliation{%
  \institution{Department of Statistics, Stanford University}
  \city{Stanford}
  \state{CA}
  \country{USA}
}

\author{Xinyue Xu}
\authornotemark[1]
\email{xinyuexu1999@gmail.com}
\affiliation{%
  \institution{College of Engineering \& Computer Science, Australian National University}
  \city{Canberra}
  \state{ACT}
  \country{Australia}
 }
 
\author{Shengyi Jiang}
\email{syjiang@cs.hku.hk}
\affiliation{%
  \institution{Department of Computer Science, The University of Hong Kong}
  \city{}
  \state{Hong Kong}
  \country{China}
 }
 
\author{Hao Wang}
\email{hoguewang@gmail.com}
\affiliation{%
  \institution{Department of Computer Science, Rutgers University}
  \city{Piscataway}
  \state{NJ}
  \country{USA}
 }
 
\author{Lu Mi}
\authornote{Corresponding author.}
\email{lumi@mit.edu}
\affiliation{%
  \institution{Computer Science and Artificial Intelligence Laboratory, Massachusetts Institute of Technology}
  \city{Cambridge}
  \state{MA}
  \country{USA}
 }

\renewcommand{\shortauthors}{et al.}
\newcommand{\methodname}{GNNS}
\begin{abstract}
  Small subgraphs (graphlets) are important features to describe fundamental units of a large network. The calculation of the subgraph frequency distributions has a wide application in multiple domains including biology and engineering. Unfortunately due to the inherent complexity of this task, most of the existing methods are computationally intensive and inefficient. In this work, we propose \methodname, a novel representational learning framework that utilizes graph neural networks to sample subgraphs efficiently for estimating their frequency distribution. Our framework includes an inference model and a generative model that learns hierarchical embeddings of nodes, subgraphs, and graph types. With the learned model and embeddings, subgraphs are sampled in a highly scalable and parallel way and the frequency distribution estimation is then performed based on these sampled subgraphs. Eventually, our methods achieve comparable accuracy and a significant speedup by three orders of magnitude compared to existing methods.
\end{abstract}



\maketitle

\section{Introduction}
Network analysis has a wide application in biology~\cite{costa2011analyzing}, chemistry~\cite{sole2004large}, engineering~\cite{milo2002network}, social science~\citep{juszczyszyn2008local} and communications~\cite{itzkovitz2005subgraphs}. Such analysis is applied in multiple aspects, including studying the distributions of nodes and edges as well as local structure (graphlets and motifs). Although graph mining has a significant impact in multiple domains, most of existing algorithms are extremely computationally intensive and only applicable to small graphs. 

In this work, we study a fundamental and important problem in network analysis, to estimate the distribution of subgraphs with the same topology structure given a target graph. Instead of counting the exact number of topology frequency~\citep{hovcevar2014combinatorial}\citep{melckenbeeck2018efficiently}\citep{ribeiro2010g}, which involves computing graph isomorphism with a time complexity of NP-complete, we use the idea of sampling to estimate the density of each subgraph type. Different from those traditional sampling-based methods include MFinder~\citep{milo2002network}, Monte Carlo Markov Chain (MCMC)-sampling~\cite{saha2015finding} etc., we propose \textbf{GNNS} (Graph Neural Network sampling), a novel, fast and scalable learning-based method using graph neural network for sampling. In particular, we utilize the idea of representation learning with a graph variational auto-encoder (VAE), to learn a correlated node embedding with graph convolutions. Then we perform a node-based embedding and extract its connected component as a subgraph. Given that a large mount of subgraphs are sampled independently and simultaneously, our algorithm is highly scalable and parallel. Eventually, our proposed algorithm could achieve a significant speed-up rate with three orders of magnitude with a comparable accuracy compared to the state-of-the-art sampling approach, e.g. MCMC-sampling.

\section{Problem Statement}

In this section, we present definitions of some fundamental concepts in graph theory that are important for formulating our problem statement. Note we only consider undirected graphs in this paper.

\begin{definition}
[\textbf{Graph}]
Denote $G(V,E)$ as a graph where V is the set of nodes and E the set of edges. Each edge $e\in E$ is denoted by $(u, v)$ where $u\in V$ and $v\in V$.
 \end{definition}

\begin{definition}
[\textbf{Subgraph}]
A graph $G'(V',E')$ is a subgraph of graph $G(V,E)$ if $V'\subseteq V$ and $E'\subseteq E$. We then denote this as $G'\subseteq G$. If $|V'| = k$, we say $G'(V',E')$ is a k-subgraph. 
\end{definition}

\begin{definition}
[\textbf{Graph Isomorphism}]
Suppose $G_1(V_1,E_1)$ and $G_2(V_2,E_2)$ are two graphs. Then $G_1(V_1,E_1)$ and $G_2(V_2,E_2)$ are isomorphic if there exists a bijection $f:V_1 \mapsto V_2$ such that $(u, v) \in E_1$ if and only if $(f(u),f(v))\in E_2$.
\end{definition}

\begin{definition}
[\textbf{Subgraph Type}]
Two subgraphs $G_1$ and $G_2$ of a graph $G$ belong to the same subgraph type $t$ if $G_1$ is isomorphic to $G_2$. 
\end{definition}

\begin{definition}
[\textbf{Frequency}~\cite{ribeiro2021survey}]
The frequency of a subgraph type $t$ in $G$ is the number of different subgraphs $G'$ of G that belong to $t$. Note when counting the frequency, two subgraphs $G'(V',E')$ and $G''(V'', E'')$ are considered different when either $V'\neq V''$ or $E'\neq E''$.
\end{definition}

We are now ready to formally define our quantity of interest:  
\begin{definition}
[\textbf{(Normalized) Subgraph Frequency Distribution}]
\label{def:distribution}
Let 
($t_1, t_2, ..., t_m$) be the list of all subgraph types in $G$ that has node number $k$. The k-subgraph frequency distribution is then \\ $(\frac{n_1}{\sum_{i=1}^{m}n_i}, \frac{n_2}{\sum_{i=1}^{m}n_i}, ......, \frac{n_m}{\sum_{i=1}^{m}n_i})$ where $n_i$ is the frequency of $t_i$ in $G$.
\end{definition}

Here are some other definitions that will be used in our methodology section~\ref{sec:methodology}:

\begin{definition}
[\textbf{Induced Subgraph}]
An induced subgraph of graph $G(V,E)$ is a subgraph $G''(V'',E'')$ where $V''\in V$ and $\forall u, v\in V''$, $(u,v)\in E''$ if and only if $(u,v)\in E$.
\end{definition}

\begin{definition}
[\textbf{Degree Distribution}]
For a graph $G(V,E)$, the degree of a node $u$, denoted as $N(u)$, is the number of nodes $v$ such that $(u,v) \in V$. The degree distribution $P(k)$ is then defined as the fraction of nodes in $G$ with degree $k$.
\end{definition}

\begin{figure*}
    \centering
    \includegraphics[width=165mm]{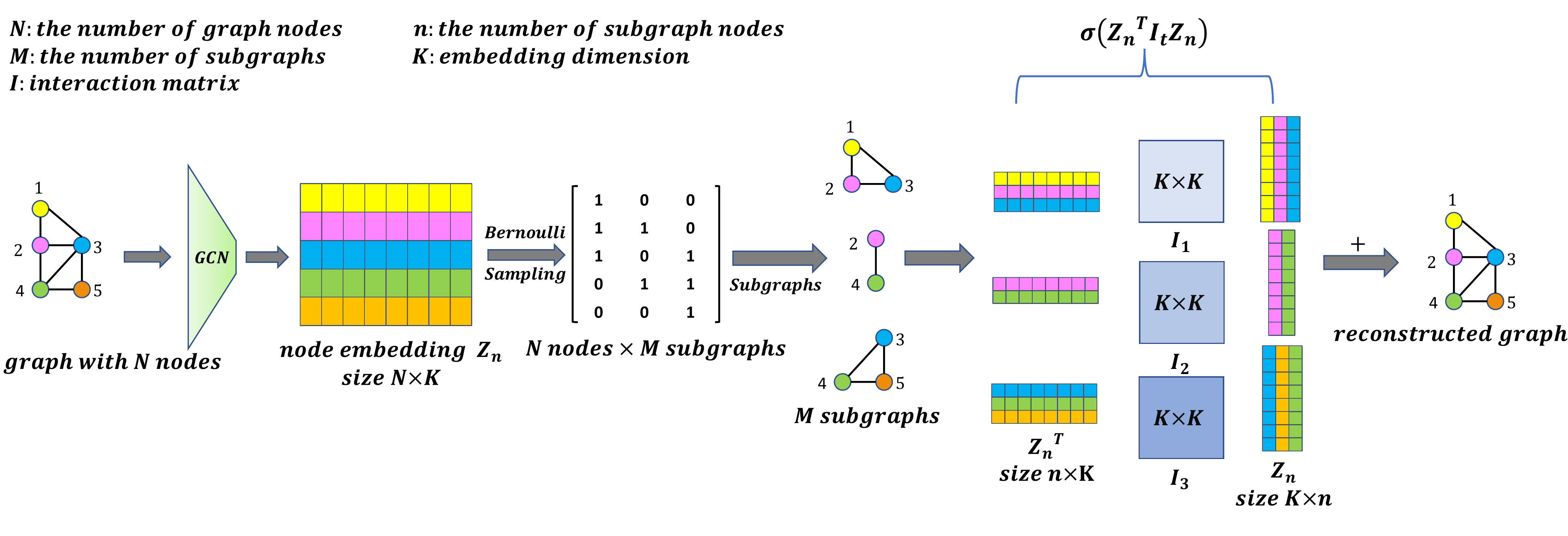}
        \vspace{-4mm}
    \caption{Overview of {\methodname}. Our method contains an inference model $q(Z|X,A)$ and a generative model $q(A|Z)$ for subgraph sampling. We utilize node embedding generated from GCN to perform node sampling, and extract the largest connected component from the sampled nodes as subgraphs. And then reconstructing the input graph with $Z_n^T I_t Z_n$ using the node embedding $Z_n$ from all subgraphs and interaction matrices $I_t$ of all subgraph types. Each node and corresponding nodes embedding are marked with the same color. The trapezoid represents the GCN layer. Different interaction matrix uses different shades of color to distinguish.}
    \label{fig:GNNS_model}
\end{figure*}

\section{Related WorkS}
Following the classification of Ribeiro~\cite{ribeiro2021survey}, the studies of subgraph density estimation can be classified into two categories, namely, exact estimation and approximate estimation. 

\subsection{Exact Methods}
Exact method consists of counting the exact number of occurrences of each subgraph type. 
Classical methods enumerate all k-nodes subgraphs before classifying them using graph isomorphism techniques such as Nauty~\cite{mckay2007nauty}. 
Milo et al. presented the MFinder~\citep{milo2002network} technique to calculate subgraphs based on the first description of Motif. Wernicke subsequently presented a new approach, FANMOD~\cite{wernicke2005faster}, to improve MFinder so that the identical subgraph is only calculated once. Kashani et al. proposed Kavosh~\cite{Kashani2009} which enhances efficiency by locating and eliminating a certain node from all subgraphs, hence minimizing the need to cache information. The isomorphic classification step required by the conventional technique frequently costs a lot of time. To improve the calculation efficiency of specific subgraph categories, the single graph search methods were devised, such as Grochow~\citep{grochow2007network}, NeMo~\cite{koskas2011nemo} and ISMAGS~\cite{demeyer2013index}. 
G-tries~\cite{ribeiro2010g} expanded the application of the enumeration method to more generic circumstances. G-tries encapsulated the topological information shared by all subgraphs in a number of subgraphs and proposed an algorithm that stores a list of subgraphs in a specialized data structure and hence counts the number of occurrences of each subgraph in the target graph efficiently. 

In addition to enumerating subgraphs explicitly, we can also count subgraphs by means of analysis. Construct a linear equation by relating the frequency of each subgraph to subgraphs with less or equal size. ORCA~\cite{hovcevar2014combinatorial} derives a system of equations that relate the orbit counts and solve for them using integer arithmetic. We use ORCA to compute the ground truth subgraph distribution in our experiment section~\ref{sec:Experiments}.
In addition to the analytical methods of linear algebra, there are decomposition methods that locate each subgraph by common adjacency, like ACC-MOTIF~\cite{meira2012accelerated}, PGD~\cite{ahmed2017graphlet} and ESCAPE~\cite{pinar2017escape}.


\subsection{Approximate Methods}
Despite the high accuracy achieved by exact estimation algorithms, 
it is inefficient when estimating the distribution of subgraphs with a large number of nodes. With the ever-increasing graph size, approximate methods are preferable to ensure computational efficiency. Randomized enumeration methods such as \citep{wernicke2005faster,ribeiro2010efficient,paredes2015rand} extends works on exact counting to perform approximation in a similar way. Another family of works is based on theories of random walk in graph:  MCMC~\cite{saha2015finding} and Guise~\cite{bhuiyan2012guise} are based on Markov chain Monte Carlo (MCMC). WRW~\cite{han2016waddling} is also a random walk based method that can derive the concentration of subgraphs of any size. Moreover, there is a family of algorithms that relies on the idea of sampling path subgraphs~\cite{seshadhri2013triadic,wang2017moss}. Lastly, a group of algorithms is based on the idea of color coding\cite{alon1995color,zhao2010subgraph,slota2013fast}.



\section{Methodology}
\label{sec:methodology}

In this section, we present our method {\methodname} to efficiently sample sub-graphs from a large graph. {\methodname} can be splitted into two major phases. We first train an auto-encoder to learn an embedding for each node. Then we perform the subgraph sampling based on the learned embedding. The sampling process can be conducted in a fully-parallel manner, which guarantees {\methodname} as a much faster and more scalable method compared with traditional methods such as MCMC.


The learning is based on the assumptions of the generation process of the graph. Specifically, we assume that a large graph consists of a random mixture of instances, where each instance is characterized by (1) a distribution over all the nodes (2) a distribution of several types. For the sampling part, we use the relaxed Bernoulli distribution \cite{maddison2016concrete} to make the sampling process differentiable and therefore ready for the back-propagation.

\subsection{Inference Model}

Given a graph $G(V,E)$ with $N$ nodes, we introduce an adjacency matrix $A$ to indicate the connectivity of the graph, and one-hot vector $X$ as node feature to indicate the identity of each node. We implement a graph variational auto-encoder that is composed of an inference model as in algorithm~\ref{alg: inference} and a generative model as in algorithm~\ref{alg: generative}.

An inference model takes an input graph with its adjacency matrix $A$ and node feature $X$ with two-layer graph convolutional network (GCN) to generate node embedding $Z_n$ with a size $N\times K$. Then subgraph embedding $Z_s$ with a size $M\times K$, is computed with node embedding $Z_n$ after subgraph sampling, indicated as $q(Z_s|Z_n)$. And then a subgraph is predicted from subgraph embedding, denoted as $q(Z_t|Z_s)$ with a size of $M\times T$. Here $M$ is the number of sampled subgraphs, $K$ is the dimension of node embedding, $T$ is total number of subgraph types.
\begin{equation}
  q(Z|X,A) = q(Z_t, Z_n, Z_s|X, A) = q(Z_t|Z_s) q(Z_s|Z_n) q(Z_n|X, A)
\end{equation}

Each node embedding ${z_n}_i$ is passed as logits into a Bernoulli distribution, and then perform node sampling $n_i \sim Bernoulli({z_n}_i)$ with a pre-defined threshold.
\begin{equation}
    q(Z_n|X, A) = \prod_{i=1}^{N}q({z_n}_i|X, A)
\end{equation}
In order to sample subgraphs, note that a naive node sampling may generate several unconnected components of the original graph (e.g. several disjoint subgraphs). To avoid such a problem, we sample the subgraph $G'(V', E')$ from its maximal connected component based on the sampled nodes $\{n_i\}$ and $\{n_j\}$ with their connected edges $\{e_{ij}\}$. Then $M$ subgraphs could be simultaneously sampled using such a procedure. The embedding ${z_s}_m$ of the subgraph $m$ is then represented as ${z_s}_m = \sum_{i \in V'} {z_n}_i$, which is a sum over all the embeddings of the nodes in this subgraph $G'$. Next, the subgraph type embedding ${z_t}_m$ is predicted using a multi-layer perceptron (MLP) from a softmax distribution given by ${z_t}_m = \text{MLP}({z_s}_m)$; one can then obtain the subgraph type $t =\argmax_t q(t|{z_t}_m)$.

\subsection{Generative Model}

Given an edge $e_{ij}$ between node $n_i$ and $n_j$ in subgraph $m$, a generative model $p(A|Z)$ uses the estimated node embedding $Z_n$ and the subgraph type $t$ predicted from subgraph embedding $Z_s$ to generate the edges $e_{ij}$ in the adjacency matrix $A$ of the input graph $G(V, E)$. In order to model for each subgraph type $t$, we introduce a trainable interaction matrix $I_t$ for each type. Given a sampled subgraph $m$, we formalize the generation of edge $p(e_{ij}) = \sigma({z_n}_i^TI_t{z_n}_j)$ of $E'$ using the learned node embedding $z_n$ and predicted subgraph type $t$. Given $M$ subgraphs with all $T$ types are sampled from an input graph at the same time, we represent the probability of reconstructed edge by the following summation over all types and subgraphs
\begin{equation}
p(A|Z)=\sum_{i=1}^N\sum_{j=1}^N\sum_{t=1}^T \sum_{m=1}^M \sigma ({z_n}_i^T I_t {z_n}_j), i\in m, j \in m
\end{equation}
\subsection{Objective Function}

We optimize our inference model $q(Z|X,A)$ and generative model $p(A|Z)$ using the evidence lower bound (ELBO), which maximizes the data likelihood and minimizes the KL divergence between the approximated posterior distribution and prior distribution. We assume a non-informative prior distribution $p(Z)$. The objective function is as follows
\begin{equation}
    \mathcal{L} = \mathbb{E}_q(Z|X,A)[\text{log} p(A|Z)] - \text{KL}(q(Z|X,A)||p(Z))
\end{equation}
\begin{algorithm}
 \begin{algorithmic}[1]

\STATE Compute node embedding $Z_n = GNN(X, A)$ from graph $G(V, E)$

\FOR{each node $i$}
\STATE Sample node using $n_i \sim Bernoulli({z_n}_i)$
\ENDFOR
\STATE Extract the largest connected component from the set of sampled nodes ${n_i}$ as a sampled subgraph $G'(V',E')\subseteq G(V,E)$  

\FOR{each subgraph $m$}
\STATE Compute subgraph embedding
${z_s}_m = \sum_{i \in m} {z_n}_i$
\STATE Compute subgraph type embedding ${z_t}_m = \text{MLP}({z_s}_m)$
\STATE Predict subgraph type $t =\argmax_t q(t|{z_t}_m)$

\ENDFOR

\caption{Inference model $q(Z|X, A)$ of \methodname}
\label{alg: inference}
\end{algorithmic}
\end{algorithm}

\begin{algorithm}
 \begin{algorithmic}[1]
 
\FOR{each subgraph type $t$}
	\FOR{each subgraph $m$}
		\FOR{each node pair $(i, j)$ inside subgraph $m$}
		\STATE generate edge $p(e_{ij}) =\sigma({z_n}_i^T I_t {z_n}_j)$
\ENDFOR
\ENDFOR
\ENDFOR
\STATE sum $p(e_{ij}), i \in m, j \in m$ over all $M$ subgraphs  and $T$ subgraph types to reconstruct the adjacency matrix $A$ from graph $G(V, E)$

\caption{Generative model $p(A|Z)$ of \methodname}
\label{alg: generative}

\end{algorithmic}
\end{algorithm}

\section{Experiments}

\begin{figure*}
    \centering
    \includegraphics[width=178mm]{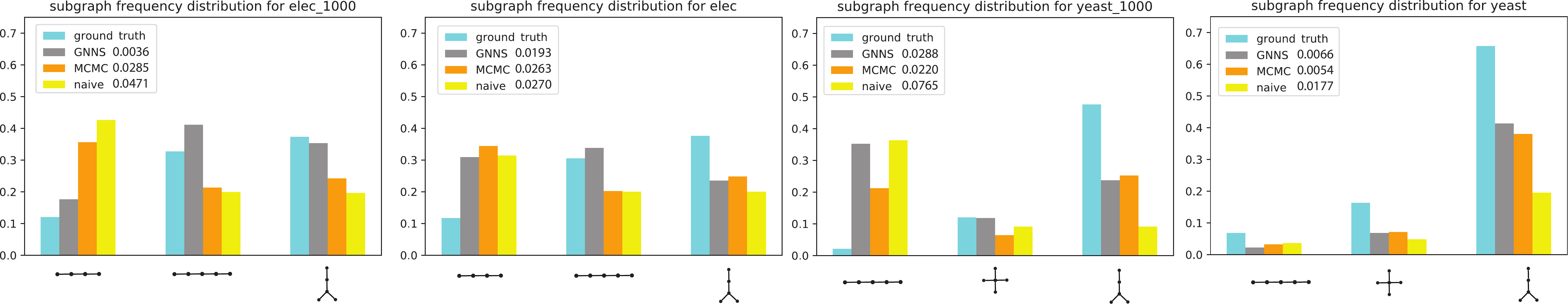}
    \vspace{-5mm}
    \caption{Approximated distributions of subgraph types with top 3 frequencies for each method. We evaluate the accuracy of the sampled distribution from our proposed method {\methodname} compared with MCMC~\cite{saha2015finding}, naive sampling and ground truth~\cite{hovcevar2014combinatorial}. Mean squared error (MSE) of the 3 most representative graphs on both Yeast and Elec dataset is provided.}
    \label{fig:GNNS}
    
\end{figure*}

We evaluate GNNS on both real world graphs and simulated graphs for subgraph types that have 4-nodes and 5-nodes. All experiments are implemented on NVIDIA GeForce RTX 2080 Ti. All training parameters are updated by the Adam Optimizer with a learning rate of $1\times 10^{-3}$. For the parameters of the model architecture, we choose the pre-defined parameters M (number of sampled subgraphs) = 1024, K (dimension of node embedding) =256 and T (total number of subgraph types) = 16. Note these parameters are based purely on empirical experience. Since most graphs have a long-tailed density distribution, underestimating subgraph types would not be an issue.
We compare the result with the MHRW version of the MCMC sampling method ~\cite{saha2015finding} where we used the authors' published code. For completeness, we also introduce a naive sampling method that enables estimation by uniformly drawing nodes from graphs to form induced subgraphs. We evaluate the methods' performance from 2 aspects: accuracy and runtime. To compute the accuracy, we obtain the ground truth subgraph frequency distribution from the exact counting method: ORCA~\cite{hovcevar2014combinatorial}. We then use the mean squared error (MSE) between the models' estimated frequency distribution and the ground truth distribution to assess the accuracy.

\label{sec:Experiments}
\subsection{Dataset}
For experiments, we select one biological network \textemdash the Yeast transcription network (with 688 nodes and 1046 edges) and one network from engineering \textemdash the Electrical network (with 252 nodes and 397 edges) \cite{Kashani2009}.

For network pre-processing, we apply the same technique as in the MCMC method: we check for undirected graphs and remove duplicate edges in the network. After that, we randomly generated 1000 random graphs following the same degree distribution as each network. We then do a  Train-Valid-Test split on the 1000 random graphs dataset. The split ratio is 8:1:1. The goal here is to accurately predict the subgraph frequency distribution the test set based on the model trained on the training set. Note the training set and test set share a same degree of distribution, which means the model should be able to generalize to the test set.

\subsection{Accuracy Comparison}
We use the mean squared error (MSE) between the ground truth distribution and the estimated distribution as the metric of estimation performance. For convenience, we combine the subgraphs with 4 and 5 nodes when computing the subgraph frequency distribution (i.e., the denominator of each term in the distribution vector in Def~\ref{def:distribution} is now the sum of the frequency of all 4-subgraph types and 5-subgraph types.) Table~\ref{tab:MSE} shows that our methods outperform MCMC (and naive sampling) in most circumstances. In Figure~\ref{fig:GNNS}, we visualize the approximated distribution of each method as a bar chart and demonstrate three subgraph types with the highest frequency.

\begin{table}
  \label{tab:metric}
  \begin{tabular}{ccccl}
    \toprule
    Dataset &{\methodname}&Naive Sampling &MCMC\\
    \midrule
    Scale & $1\mathrm{e}{-3}$ & $1\mathrm{e}{-3}$ & $1\mathrm{e}{-3}$ \\
    \midrule
    Elec & \textbf{2.32}& 3.64& 3.30\\
    Yeast & \textbf{6.55}& 17.20& 7.85\\
    Elec1000 & \textbf{0.73}& 3.98& 3.50\\
    Yeast1000 & 8.83& 13.30& \textbf{5.32}\\
  \bottomrule
\end{tabular}
\caption{Sampling quality comparsion (MSE). Elec and yeast stand for the single original graphs. Elec1000 and Yeast1000 stand for the average of 1000 random graphs sampled from Elec and Yeast, respectively.}
\label{tab:MSE}
\end{table}

\subsection{Runtime Comparison}
We use the total running time on the 1000 random graphs in the three approaches for runtime comparison. Table~\ref{tab:freq} shows the results. GNNS tallied the forward and sampling procedure, whereas Naive Sampling and MCMC only tallied the sampling time. GNNS is clearly superior to MCMC in terms of time, which significantly enhances the computational performance. The acceleration of deep learning demonstrates the significant potential of our model for estimating the frequency distribution of subgraphs.

\begin{table}
  \begin{tabular}{cccl}
    \toprule
    Dataset &{\methodname}&Naive Sampling &MCMC\\
    \midrule
    Scale & $1\mathrm{e}{-2}$ & $1\mathrm{e}{-2}$ & $1\mathrm{e}{-2}$ \\
    \midrule
    Elec1000 &  12.01  & 0.81  & 58400 \\
    Yeast1000 & 24.45  & 0.86  & 257100 \\
  \bottomrule
\end{tabular}
\caption{Sampling time comparison (in seconds). Elec1000 and Yeast1000 stand for the running time on the 1000 random graphs sampled from Elec and Yeast, respectively.}
\label{tab:freq}
\end{table}

\section{Conclusion}

In this work, we propose GNNS, a learning-based representation framework, which utilizes graph neural networks to learn hierarchical embeddings on the node, subgraph, and type levels. 
We perform subgraph sampling in a highly scalable and parallel way. Experiments show that our proposed framework achieves a comparable accuracy and a significant speed-up compared to MCMC-sampling with three orders of magnitude. We also think of a lot of interesting future directions, e.g. motif search, that could benefit from our proposed graph neural network based sampling. \\

\bibliographystyle{ACM-Reference-Format}
\bibliography{sample-base}


\end{document}